\documentclass[10pt,twocolumn,letterpaper]{article}

\usepackage[final]{cvpr}

\usepackage{times}
\usepackage{epsfig}
\usepackage{graphicx}
\usepackage{amsmath}
\usepackage{amssymb}

\usepackage[utf8]{inputenc}
\usepackage{multirow}
\usepackage{rotating}
\usepackage{verbatim}
\usepackage{booktabs}
\usepackage{array}
\usepackage{textcomp, gensymb}
\usepackage{diagbox}
\usepackage[T1]{fontenc}    
\usepackage{url}            
\usepackage{booktabs}       
\usepackage{amsfonts}       
\usepackage{nicefrac}       
\usepackage{microtype}      
\usepackage{setspace}
\usepackage{multirow}
\usepackage{amssymb}
\usepackage{diagbox}
\usepackage{mathtools}
\usepackage{wrapfig}
\usepackage{diagbox} 
\usepackage{wrapfig,lipsum,booktabs}
\usepackage{caption}
\usepackage{appendix}

\usepackage{color}
\usepackage{xcolor}
\definecolor{citecolor}{rgb}{0.21,0.49,0.74}
\usepackage{colortbl}
\usepackage{array}

\usepackage[pagebackref=false,breaklinks=true,letterpaper=true,citecolor=citecolor,colorlinks,bookmarks=false]{hyperref}

\definecolor{tblue}{RGB}{80,80,245}
\definecolor{tred}{RGB}{250,100,100}

\newcolumntype{x}[1]{>{\centering\arraybackslash}p{#1pt}}

\newcommand{\app}{\raise.17ex\hbox{$\scriptstyle\sim$}}

\newlength\savewidth\newcommand\shline{\noalign{\global\savewidth\arrayrulewidth
  \global\arrayrulewidth 1pt}\hline\noalign{\global\arrayrulewidth\savewidth}}
\newcommand{\tablestyle}[2]{\setlength{\tabcolsep}{#1}\renewcommand{\arraystretch}{#2}\centering\footnotesize}

\newcommand{\padcell}{\cellcolor{citecolor!10}}
\newcommand{\padcellred}{\cellcolor{red!10}}

\usepackage{amssymb}
\usepackage{pifont}

\newcommand{\cmark}{\ding{51}}%
\newcommand{\xmark}{\ding{55}}%
\newcommand{\modelname}[0]{\mbox{\textsc{FORGE}}\xspace}

\def\paperID{107}
\def\confName{3DV}
\def\confYear{2024}

\begin{document}

\title{Few-View Object Reconstruction with Unknown Categories and Camera Poses}

\author{Hanwen Jiang \quad Zhenyu Jiang \quad Kristen Grauman \quad Yuke Zhu \\[2mm]
 Department of Computer Science, The University of Texas at Austin \\{\tt\small \hspace{0mm}\{hwjiang,zhenyu,grauman,yukez\}@cs.utexas.edu}
}
\maketitle

\begin{abstract}
While object reconstruction has made great strides in recent years, current methods typically require densely captured images and/or known camera poses, and generalize poorly to novel object categories. 
To step toward object reconstruction in the wild, this work explores reconstructing general real-world objects from a few images without known camera poses or object categories. 
The crux of our work is solving two fundamental 3D vision problems --- shape reconstruction and pose estimation --- in a unified approach. Our approach captures the synergies of these two problems: reliable camera pose estimation gives rise to accurate shape reconstruction, and the accurate reconstruction, in turn, induces robust correlations between different views and facilitates pose estimation.
Our method \modelname{} predicts 3D features from each view and leverages them in conjunction with the input images to establish cross-view correlations for estimating relative camera poses. The 3D features are then transformed by the estimated poses into a shared space and are fused into a neural radiance field.
The reconstruction results are rendered by volume rendering techniques, enabling us to train the model without 3D shape ground-truth. Our experiments on both real and synthetic datasets, as well as in the wild images, show that \modelname{} reliably reconstructs objects from five views. Our pose estimation method outperforms existing ones by a large margin. The reconstruction results under predicted poses are comparable to the ones using ground-truth poses. And the performance on novel testing categories matches the results on categories seen during training. 
\end{abstract}

\vspace{-0.1in}
\section{Introduction}
Reconstructing real-world objects through the lens of RGB cameras is crucial in AR/VR applications~\cite{bekele2018survey,rahaman2019photo}, embodied AI~\cite{szot2021habitat,habitat19iccv} and robotics~\cite{jiang2021synergies,yan2018learning}. Conventional methods~\cite{Schnberger2016StructurefromMotionR, Klein2009ParallelTA, Moulon2016OpenMVGOM} rely on densely captured images for optimization-based reconstruction. 
The stringent requirement on dense inputs hinders the broad applicability of these methods. 
In contrast, few-view object reconstruction~\cite{Yang2022FvOR,Yu2021pixelNeRFNR} aims to quickly create 3D models from a few images of real-world objects, such as online product snapshots. 
Yet, existing work requires either 3D ground-truth supervision for training~\cite{Choy20163DR2N2AU} or well-calibrated camera poses for inference~\cite{Yu2021pixelNeRFNR, Kar2017LearningAM}, and often restricts to seen object categories~\cite{Yang2022FvOR}. To move toward efficient capture of real-life objects, we aspire to develop a practical approach that performs general few-view object reconstruction with no reliance on object categories or camera poses.

\begin{figure}[t]
\centering
\includegraphics[width=0.97\linewidth]{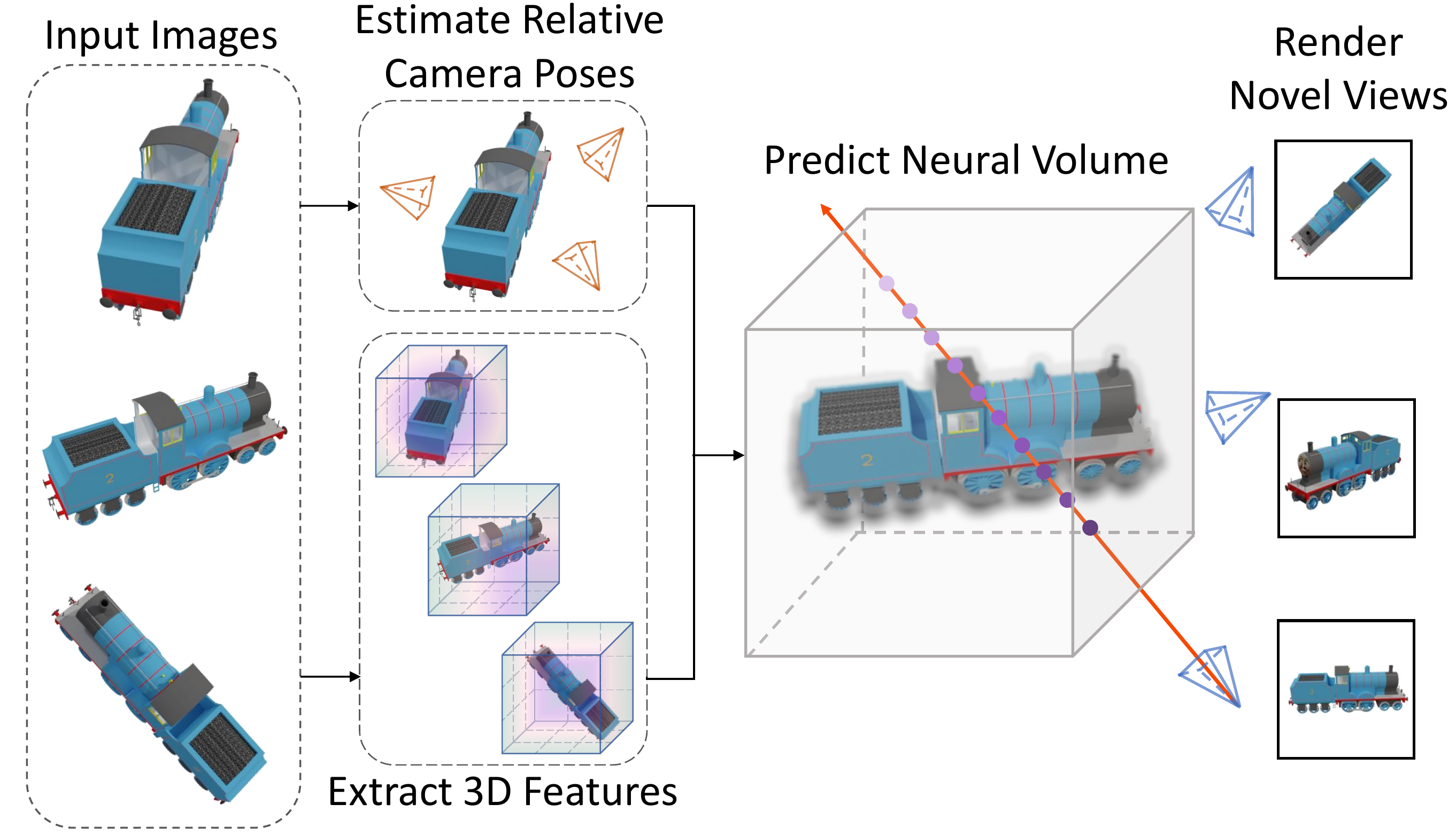}
\vspace{-0.12in}
\caption{\small{\textbf{Few-view object reconstruction.} \modelname{} reconstructs a 3D object from a few views without camera poses. It estimates relative camera poses between input views and extracts per-view 3D features. The features are fused based on the predicted poses to predict a neural volume, encoding the radiance field. We use volume rendering to generate the reconstruction results.}
}
\vspace{-0.2in}
\label{fig: teaser}
\end{figure}

In this paper, we introduce \modelname{} (\underline{F}ew-view \underline{O}bject \underline{R}econstruction that \underline{GE}neralizes), illustrated in Fig.~\ref{fig: teaser}. Departing from category-level methods that reconstruct objects in a category-specific canonical space~\cite{Wang2019NormalizedOC,Kanazawa2018LearningCM},  \modelname{} encodes individual inputs into 3D features in their own camera spaces. It then estimates relative camera poses between input views and transforms the 3D features into a shared reconstruction space with the estimated poses. This design eliminates the need for a canonical reconstruction space of each category and enables \modelname{} to generalize across object categories. The transformed 3D features are subsequently aggregated into a neural volume. We follow NeRF\cite{Mildenhall2020NeRFRS} and use differentiable volume rendering to predict novel views from the neural volume. The model is supervised with a reconstruction loss between the rendered and raw input images, without 3D supervision for object shape.

The foremost challenge is estimating relative camera poses between views. Existing works on camera pose estimation leverage correlations of 2D images~\cite{schoenberger2016sfm, lowe2004distinctive, zhang2022relpose}. For few-view object reconstruction, drastic variation of camera poses impedes the establishment of robust 2D correlation. To this end, we design a novel relative pose estimator with both 3D features and 2D images as input, exploiting the correlation between the 3D features to eliminate reprojection ambiguity. Furthermore, to avoid compounding errors in pairwise relative pose estimation~\cite{Lai2021VideoAS, Rockwell2022}, \modelname{} predicts poses based on the correlations across all inputs. It thus benefits from the 3D correlation and has a global understanding of the camera configurations.


\modelname{} exploits the synergies between shape reconstruction and pose estimation to improve performance on both. 
We first train the model with ground-truth camera poses. Using ground-truth camera poses encourages the model to learn 3D geometry priors for extracting consistent 3D features from each input view. We then learn the relative camera pose estimator, which builds 3D correlation accurately in the well-established 
view-consistent 3D feature space.

\begin{table}[t]
\scriptsize
\tablestyle{5pt}{1.05}
\setlength\tabcolsep{3pt}
\begin{tabular}{l|cccccc}
 & \multicolumn{1}{c}{(1)}
 & \multicolumn{1}{c}{(2)}
 & \multicolumn{1}{c}{(3)}
 & \multicolumn{1}{c}{(4)}
 & \multicolumn{1}{c}{(5)}
 & \multicolumn{1}{c}{(6)}
 \\
 & \multicolumn{1}{c}{Sparse}
 & \multicolumn{1}{c}{W/o}
 & \multicolumn{1}{c}{Novel}
 & \multicolumn{1}{c}{Feed-}
 & \multicolumn{1}{c}{W/o}
 & \multicolumn{1}{c}{Explicit}
 \\
 & \multicolumn{1}{c}{views}
 & \multicolumn{1}{c}{poses}
 & \multicolumn{1}{c}{category}
 & \multicolumn{1}{c}{forward}
 & \multicolumn{1}{c}{shape}
 & \multicolumn{1}{c}{Recon.}
 \\
\shline
NeRF~\cite{Mildenhall2020NeRFRS} & \padcellred \xmark & \padcellred \xmark & \padcell \cmark &\padcellred \xmark & \padcell \cmark & \padcell \cmark  \\
RegNeRF~\cite{Niemeyer2021RegNeRFRN} & \padcell \cmark & \padcellred \xmark & \padcell \cmark & \padcellred \xmark & \padcell \cmark & \padcell \cmark\\
P-NeRF~\cite{Yu2021pixelNeRFNR} & \padcell \cmark & \padcellred \xmark & \padcell \cmark & \padcell \cmark & \padcell \cmark & \padcell \cmark \\
SRT~\cite{srt22} & \padcell \cmark & \padcell \cmark & \padcell \cmark & \padcell \cmark & \padcell \cmark & \padcellred \xmark \\
FvOR~\cite{Yang2022FvOR} & \padcell \cmark & \padcell \cmark & \padcellred \xmark &  \padcellred \xmark & \padcellred \xmark & \padcell \cmark \\
FORGE & \padcell \cmark & \padcell \cmark & \padcell \cmark & \padcell \cmark & \padcell \cmark & \padcell \cmark
 \\ 
\end{tabular}
\vspace{-0.1in}
\caption{\small{\textbf{Model feature comparison.} (1) Model operates on a few sparse input views; (2) Ability to infer without camera poses of inputs; (3) Ability to perform reconstruction on novel category; (4) Fast, feed-forward reconstruction; (5) No need for 3D ground-truth of object shape; (6) Ability to produce explicit 3D reconstruction of point cloud, mesh or voxel.}}
\label{tabel: setting}
\vspace{-0.2in}
\end{table}

To evaluate the generalization ability of \modelname{}, we design a new dataset with diverse object categories and harder camera settings compared to previous datasets~\cite{Sitzmann2019SceneRN, Xiang2014BeyondPA}. We also evaluate \modelname{} on diverse real-object datasets~\cite{omniobject3d, Downs2022GoogleSO}.
Our results show that \modelname{} outperforms prior art on both reconstruction and relative camera pose estimation by a large margin. The reconstruction quality under predicted poses matches the performance using ground-truth relative camera poses. Moreover, \modelname{} exhibits a strong generalization ability on novel object categories, almost on par with the performance on seen objects.

We highlight our contributions as follows: i) We develop a few-view object reconstruction approach that jointly estimates camera poses of input images and reconstructs the objects, which generalizes to novel object categories; ii) we propose a novel camera pose estimation method to handle large camera variations of few-view inputs; iii) Our experiments demonstrate the synergies between reconstruction and pose estimation for improved quality on both tasks. 

\begin{figure*}[t]
\centering
\includegraphics[width=\linewidth]{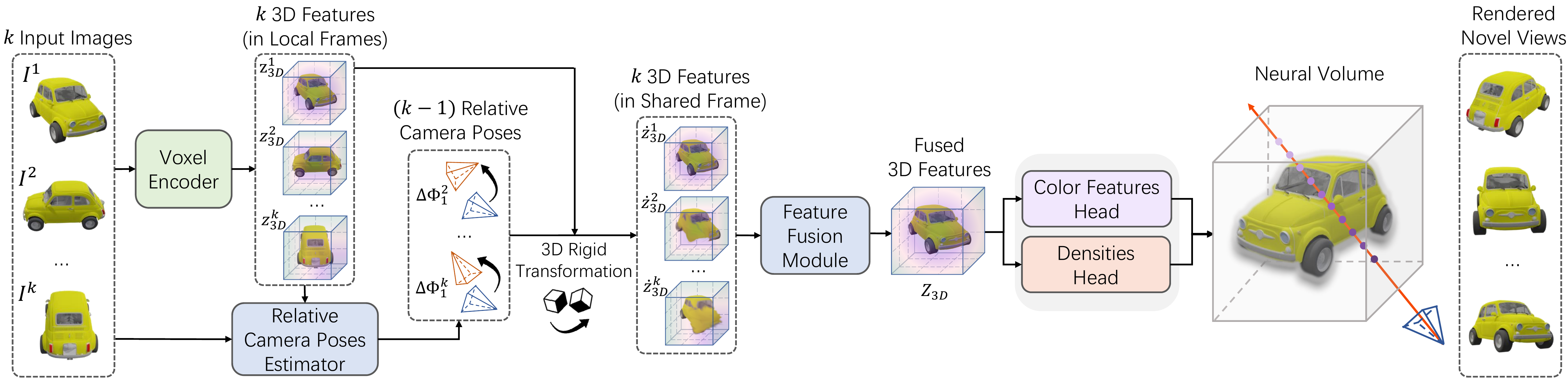}
\vspace{-0.2in}
\caption{\small{\textbf{Model overview.} \modelname{} extracts 3D voxel features from each view and uses the 3D features, along with 2D observations, to estimate relative camera poses. The 3D features in their corresponding cameras' frames are transformed into a shared reconstruction space using the rigid transformation computed by the relative camera poses. The features are fused to predict a neural volume that encodes the radiance field. We use volume rendering techniques to render the reconstruction results. }
}
\vspace{-0.15in}
\label{fig: overview}
\end{figure*}

\vspace{-0.05in}
\section{Related Work}
\vspace{-0.05in}
\paragraph{Multi-View Reconstruction.}
Reconstructing objects and scenes from multi-view images has been a long-standing problem in computer vision~\cite{Seitz2006ACA}.  Traditional methods, e.g., COLMAP~\cite{Schnberger2016StructurefromMotionR}, and their learning-based counterparts, e.g., DeepV2D~\cite{Teed2020DeepV2DVT}, have already shown great success. However, these SLAM~\cite{Klein2009ParallelTA, Teed2020DeepV2DVT, Bloesch2018CodeSLAML, Zhu2021NICESLAMNI}, SfM~\cite{Schnberger2016StructurefromMotionR, Tang2018BANetDB, Teed2020DeepV2DVT} and RGB-D registration~\cite{ElBanani2021UnsupervisedRRUP} approaches require dense-view inputs and smooth camera movement.

Another line of work aims at using only sparse or even a few views as input, where two main streams emerged. The first genre is pose-free. For example, 3D-R2N2~\cite{Choy20163DR2N2AU} and Pix2vox++\cite{Xie2020Pix2VoxMC} aggregate 3D information from all inputs directly. SRT~\cite{srt22} builds a geometry-free method using a large transformer model.
However, geometry-free methods are hard to generalize to unseen categories due to the absence of operations that aware of cross-view correlation. The other approach is pose-aware, where camera poses are used for aligning features of each view into a common reconstruction space~\cite{Kar2017LearningAM, Oechsle2021UNISURFUN, Wang2021NeuSLN, Niemeyer2020DifferentiableVR, Park2020LatentFusionED}, and they benefit from using ground-truth camera poses. In this paper, we develop a pose-aware method that predicts both shape and relative camera poses in a unified manner without using ground-truth camera poses during inference.

\vspace{-0.17in}
\paragraph{Volumetric 3D and Neural Radiance Fields.}
3D explicit volumetric representations, especially voxel grids, have been widely used for modeling objects~\cite{Girdhar2016LearningAP,Choy20163DR2N2AU, Kar2017LearningAM, Yan2016PerspectiveTN, Tulsiani2019MultiviewSF, NguyenPhuoc2019HoloGANUL, Riegler2017OctNetLD} and scenes~\cite{Tulsiani2018FactoringSP, Kulkarni20193DRelNetJO}. In recent years, given the impressive results of implicit representations on 3D vision tasks~\cite{Mescheder2019OccupancyNL, Park2019DeepSDFLC}, NeRF~\cite{Mildenhall2020NeRFRS} adopts an implicit neural radiance field for 3D volumetric representations. NeRF and its variants achieve a solid ability to model complex geometry and appearance~\cite{MartinBrualla2021NeRFIT,Park2021NerfiesDN, Schwarz2020GRAFGR, Zhang2020NeRFAA, Pumarola2021DNeRFNR}.
Nonetheless, NeRF uses one global MLP for fitting each scene, which is challenging to optimize and suffers from limited generalization.

Two types of methods are proposed to solve the limitations of conventional NeRF methods. The first is integrating 2D image features into the MLP~\cite{Yu2021pixelNeRFNR, Wang2021IBRNetLM, Reizenstein2021CommonOI}, which 
is sensitive to camera pose error due to the 3D-2D projection. The second is using a semi-implicit radiance field, where the radiance field is attached to voxel grids~\cite{Liu2020NeuralSV}. Furthermore, to make the radiance field more generalizable, some work~\cite{Chen2021MVSNeRFFG, Ye2021ShelfSupervisedMP, Zhang2022NeRFusionFR} trained 3D encoders which directly predict voxel-based radiance fields from images. However, MVSNeRF~\cite{Chen2021MVSNeRFFG} is set to a fixed number of nearby views, ShelfSup~\cite{Ye2021ShelfSupervisedMP} predicts radiance fields in the canonical space, and NeRFusion~\cite{Zhang2022NeRFusionFR} requires ground-truth camera poses. In contrast, our model predicts the voxel-based radiance field from raw images, which can be fused using an arbitrary number of views using predicted poses.

\vspace{-0.17in}
\paragraph{Reconstruction from Images without Poses.}
One drawback of pose-aware reconstruction methods using a volumetric representation is the requirement for accurate camera poses. Many works assume access to ground-truth camera poses~\cite{Mildenhall2020NeRFRS, Sitzmann2019SceneRN, Kar2017LearningAM}, which limits their applicability.
BARF~\cite{Lin2021BARFBN} and NeRS~\cite{Zhang2021NeRSNR} performed joint optimization on shape and pose. However, they still rely on highly accurate initial poses.
FvOR~\cite{Yang2022FvOR} proposed a pose initialization module but it requires 3D shape supervision.

Another line of work~\cite{Tung2019LearningSC, NguyenPhuoc2019HoloGANUL, Lai2021VideoAS} takes advantage of the synergies between shape and pose. GRNN~\cite{Tung2019LearningSC} trains a relative pose estimator and deploys it for predicting the poses during reconstruction. VideoAE~\cite{Lai2021VideoAS} learns the relative camera poses and 3D representations under fully unsupervised learning by disentangling the shape and pose. However, these two works predict the relative camera poses from raw 2D views, making them hard to handle unseen categories, texture-less objects, and large pose variations due to the 2D ambiguity.
Our method predicts relative camera poses in world coordinates. We use the predicted poses for performing cross-view fusion, linking shape and pose prediction with awareness of the underlying 3D shapes.
\section{Overview}
\vspace{-0.05in}

We study the problem of object reconstruction from a few RGB images without camera poses and category information. As Fig.~\ref{fig: overview} illustrates, our model \modelname{} learns camera pose estimation from 2D images and 3D features extracted from them. To deal with the large pose variation between the few views, our pose estimator jointly predicts the relative camera poses of all input images rather than chaining up pair-wise pose estimations. To handle objects from novel categories, we avoid learning category-specific priors. \modelname{} extracts per-view 3D features in their own camera spaces and transforms them into a shared reconstruction space with the estimated relative camera poses. Therefore we get rid of the category-specific canonical space for reconstruction. In the shared reconstruction space, we use a feature fusion module to aggregate information from single-view 3D features. Then a decoder head predicts a neural volume from the fused features. We use volume rendering techniques to produce the reconstruction results. During training, we render the results of input views and use 2D-based rendering loss as supervision. \modelname{} can be trained without any 3D supervision of object geometry. We design a new loss function for learning consistent 3D features across views. Besides, we can obtain accurate voxel reconstruction from the neural volume by a simple threshold. We introduce details of \modelname{} and the training protocol in the following.


\begin{figure*}[t]
\centering
\includegraphics[width=17.5cm]{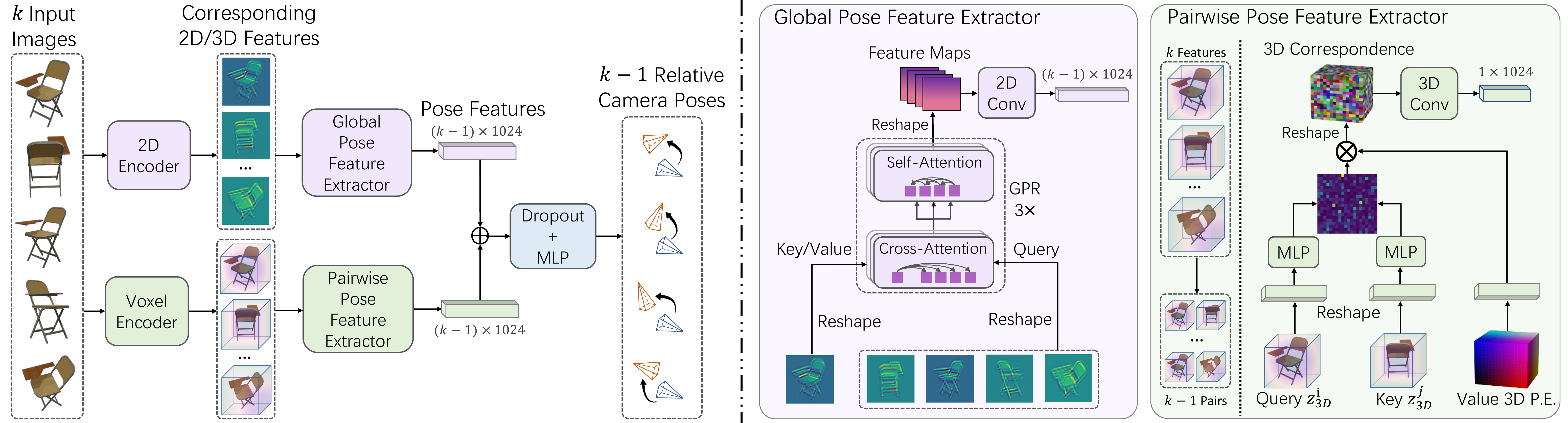}
\caption{\small{\textbf{Relative camera poses estimator.} \modelname{} uses a Global Pose Features Extractor to jointly reason about pose cues for all frames. \modelname{} also uses a Local Pose Features Extractor to predict pairwise pose features by explicitly reasoning about 3D correlation.}
}
\vspace{-0.15in}
\label{fig: pose estimation}
\end{figure*}



\section{Method}
\vspace{-0.05in}
\label{sec: method}

Here we introduce the design of \modelname{}. The input of the model is $k$ image observations $\mathcal{I} = \{I^i \vert i=1,...,k\}$ captured by corresponding cameras $\mathcal{C} = \{C^i \vert i=1,...,k\}$. The model predicts: i) 3D features $\mathcal{Z}_{3D} = \{z_{3D}^i \vert i=1,...,k\}$ of all views; ii) relative camera poses $\Delta\Phi_i = \{\Delta\Phi_i^j \vert j=1,...,k; j\ne i\}$, $\Delta\Phi_1^j \in \mathbb{SE}(3)$, where $\Delta\Phi_i^j$ is the pose of camera $C^j$ relative to camera $C^i$, and the $i^{th}$ frame is set as the canonical frame. Then the features are transformed by the relative camera poses into camera $C^i$'s frame and fused to predict the neural radiance field $V$. We use the first frame as the canonical frame by default, if not otherwise specified. We elaborate on each component in the following.

\subsection{Voxel Encoder}
\label{sec: 3d encoder}
For a view $I^i$, the encoder $F_{3D}$ encodes it into 3D voxel features $z_{3D} = F_{3D}(I)$, where $z_{3D}^i \in \mathbb{R}^{c\times d\times h\times w}$. 
In detail, we use ResNet-50~\cite{He2016DeepRL} to extract 2D feature maps $z_{2D}^i = F_{2D}(I)$, where $z_{2D}^i \in \mathbb{R}^{C\times h\times w}$.
Then we reshape $z_{2D}^i$ to 3D voxel features in $\mathbb{R}^{(C/d)\times d\times h\times w}$ as a deprojection operation. We finally perform one 3D convolution on it for refinement, which changes the 3D channel size to $c$.

\subsection{Relative Camera Pose Estimator}

Estimating relative camera poses can be an ill-posed problem under large pose variation --- the shared visible object part between two views can be small, making it hard to establish cross-view correlations.
Moreover, building 2D correlation on 2D images or feature maps~\cite{schoenberger2016sfm,lowe2004distinctive} is vulnerable under the 3D-2D projection ambiguity.

We introduce two novel pose feature extractors complementary to each other. As shown in Fig.~\ref{fig: pose estimation}, we use a global pose feature extractor, taking all 2D views as input and jointly reasoning about poses for all frames. We also build a pairwise pose feature extractor that computes explicit 3D correlation on the predicted 3D features. 
The design benefits pose estimation in three ways: i) The global pose feature extractor allows information to pass between all frames. It leverages information from other views, which may have a larger shared visible part with the canonical view, to infer the relative pose of a query view. Besides, reasoning about poses globally in 2D is much cheaper than on 3D features;
ii) Finding 3D correlation on the predicted 3D features avoids the 3D-2D ambiguity problem and makes pose estimation more accurate; and iii) The multi-modal inputs (both 2D views and 3D features) make pose estimation more robust. 

\vspace{-0.1in}
\paragraph{Global Pose Feature Extractor.} 
The global pose feature extractor takes in all views and jointly predicts pose features, denoted as $p_g \in \mathbb{R}^{(k-1)\times 1024} $, corresponding to all $k-1$ query views $\mathcal{I}_q$.
We use another 2D backbone to extract 2D feature maps of each view, denoted as $\mathbb{Z}_{2D}' = \{z_{2D}'^i \vert i=1,...,k\}, z_{2D}'^i \in \mathbb{R}^{h'\times w'\times 1024}$. We then reshape the canonical view features into a 1D vector $k_g \in \mathbb{R}^{N_{2D}\times 1024}$, where $N_{2D} = h'\cdot w'$. We similarly reshape the $k-1$ query view features as $q_g \in \mathbb{R}^{N_q\times 1024}$, where $N_q = (k-1)\cdot N_{2D}$. Then we use multiple global pose feature reasoning (GPR) modules to infer pose features. Specifically, each GPR module includes two standard multi-head transformer~\cite{Vaswani2017AttentionIA} blocks.
In the first transformer block, we perform cross-attention, where the query is the feature $q_g$ of the query view, and the key and value are features $k_g$ of the canonical view. The block reasons 2D correlation between each query view and the canonical view. Then we perform self-attention on the updated query view features. It jointly refines the correlation cues for all query views by referring information from each other. 
After the GPR modules, we reshape the updated query view features back into 2D and use 2D convolutions to down-scale the 2D resolution to $1$ to get the global pose features $p_g$.


\vspace{-0.1in}
\paragraph{Pairwise Pose Feature Extractor.}
The pairwise pose feature extractor builds $k-1$ feature pairs, where each pair consists of 3D features from a query view and the canonical view. Then it predicts pose features for them separately.
Take the query view $I^i$ and the canonical view $I^1$ as an example. The inputs are 3D voxel features $z_{3D}^i, z_{3D}^1$, and the output is the relative pose features $p_l^i \in \mathbb{R}^{1\times 1024}$. 

Specifically, we reshape the 3D features into a 1-D vector in $\mathbb{R}^{N_{3D}\times c}$, where $N_{3D} = d\cdot h\cdot w$. Then we compute the similarity tensor $S^{i}=z_{3D}^i \cdot (z_{3D}^1)^T$, where $(\cdot)^T$ is the transpose operation. We get the correlation as
$Corr_{1D}^{i} = S^{i} \cdot PE_{1D}$, where $PE_{1D}$ is the expanded high-dimensional 3D positional embedding~\cite{Vaswani2017AttentionIA}. The positional embedding represents the location of each voxel in a high-dimensional space. $PE \in \mathbb{R}^{d\times h\times w\times c}$ and $Corr_{1D}^{i} \in \mathbb{R}^{N_{3D}\times c}$. We reshape the correlation back into a 3D volume $Corr^{i}\in \mathbb{R}^{d\times h\times w\times c}$.
For each voxel of the volume, its features represent the location of its corresponding voxel location within the canonical view 3D features in high-dimensional space.
In the end, we use 3D convolutions to down-scale it to resolution $1$ to get the relative pose features $p_l^i$ of $\{I^i, I^1\}$. We concatenate the features of all query views as the final output $p_l \in \mathbb{R}^{(k-1)\times 1024}$.

\vspace{-0.15in}
\paragraph{Pose Prediction.}
We concatenate the pose features predicted by the two feature extractors to get the final features $p$.
We use an MLP to regress the relative camera poses $\Delta\Phi_i = \{\Delta\Phi_1^j \vert j=2,...,k\}$ for $k-1$ query views.
To prevent the model from overfitting to the features extracted by either one, we use a Dropout layer before the MLP with a probability of $0.6$ during training.
\subsection{Feature Fusion Module}
\label{sec: fusion}
\paragraph{Feature Transformation.} In general, we transform a 3D point in camera $C^i$'s frame into camera $C^j$'s frame by using rigid transformation $T^j_i = \Phi^j \cdot (\Phi^i)^{-1}$, where $(\cdot)^{-1}$ is the inverse operation and $\Phi$ is camera pose. Given the extracted 3D features of $k-1$ query views, we use the camera poses to transform them into the canonical view's camera frame. Specifically, the camera poses of the query views are computed by $\Phi^i = \Phi^1 \cdot \Delta \Phi_1^i$, where $\Phi_1$ is a fixed canonical pose. We denote the 3D features after transformation as $\mathbb{\dot{Z}}_{3D} = \{\dot{z}_{3D}^i \vert i=1,...,k\}$.

\vspace{-0.17in}
\paragraph{View Fusion.} 
For the transformed 3D features $\mathbb{\dot{Z}}_{3D}$, we first perform average pooling over them. The pooled 3D features serve as an initialization for sequentially fusing per-view features into final 3D features $Z_{3D}$ using ConvGRU~\cite{Ballas2016DelvingDI}.
Specifically, the fusion sequence is determined by the relative camera poses, where we first fuse features of the views closer to the canonical view. 
When the predicted camera poses are noisy, the average pooling operation keeps the low-frequency signals, and the sequential fusion recovers high-frequency details.

\subsection{Neural Volume-based Rendering}
\label{sec: decoder}

Inspired by NeRF~\cite{Mildenhall2020NeRFRS}, we reconstruct objects with radiance fields. These fields, represented as neural volumes,  are predicted in a fully \textit{feed-forward} manner.

\vspace{-0.17in}
\paragraph{Definition and Prediction.} We denote the neural volume as $V \coloneqq (V_\sigma, V_f)$, where $V_\sigma$ and $V_f$ are the density and neural feature volumes. They help reconstruct the geometry and appearance of the object.
Given the 3D feature volume $Z_{3D}$ after fusion, we use two prediction heads composed of several 3D convolution layers to predict the $V_\sigma$ and $V_f$.

\vspace{-0.17in}
\paragraph{Volume Rendering.} We read out the predicted neural volume on 2D using volume rendering as $(\hat{I}, \hat{I}_\sigma) = \pi(V,\Phi)$, where $\Phi$ is the camera pose, $\hat{I}$ and $\hat{I}_\sigma$ are rendered image and mask. We follow the volume rendering techniques in NeRF~\cite{Mildenhall2020NeRFRS}. Differently, for each 3D query point $p$, we get its 3D features by interpolating the neighbor voxel grids. 
Besides, we first render a feature map and then use several 2D convolutions to predict the final RGB image~\cite{Niemeyer2021GIRAFFERS}. 

\subsection{Training Protocol}
\label{sec: train}

We train \modelname{} in three stages. First, we train the voxel encoder (Sec.~\ref{sec: 3d encoder}), fusion module (Sec.~\ref{sec: fusion}), and volume renderer (Sec.~\ref{sec: decoder}) under ground-truth poses using $L_{3D} = L_{mv} + L_{corr}$. Specifically, $L_{mv}$ is the 2D photometric loss applied on all input views, and $L_p$ is the Perceptual loss~\cite{Johnson2016PerceptualLF}.
\begin{align}
    L_{2D} &= \vert\vert I_\sigma^i - \hat{I}_\sigma^i \vert\vert + \lambda_{img} \vert\vert I^i - \hat{I}^i\vert\vert, \\
    L_{mv} &= \frac{1}{k} \Sigma_{i=1}^{k} 
(L_{2D}(I^i_\sigma, \hat{I}^i_\sigma, I^i, \hat{I}^i) + \lambda_{p}L_p(I^i, \hat{I}^i) ),
\end{align}
The cross-view consistency loss $L_{corr}$ is designed to encourage the features that correspond to the same object part while extracted from a different view to be close in the feature space, which makes the reconstruction result more coherent. The well-established feature space also benefits following correlation-based pose estimation, as finding cross-view correlation using our similarity-based method becomes more effortless in the feature space.
We compute the loss on rendering results. We separate the input views into two sets: we use one set ($n$ views) for building the neural volume and use the cameras of the other set ($k-n$ views) to render the results and vice versa. The reconstruction of views in the second set should be reasonable.
\begin{align}
    L_{corr} = \frac{1}{k-n} \Sigma_{i=n+1}^{k} 
L_{2D}(I^i_\sigma, \ddot{I}^i_\sigma, I^i, \ddot{I}^i),
\end{align}
where $\ddot{I}$ and $\ddot{I}_\sigma$ are results rendered using a subset of views.

Second, we train the relative camera pose estimator with a loss $L_{pose} = \vert\vert \Phi - \hat{\Phi}\vert\vert^2$, where $\Phi$, $\hat{\Phi}$ are ground-truth and predicted relative camera poses. The two pose feature extractors are trained separately and then fine-tuned together. We use the quaternion representation of rotation, and compute the loss for translation and rotation separately.

Finally, we fine-tune the model end-to-end with all losses above. We empirically observe that single-stage training leads to collapse. We conjecture that the pose estimator relies on representations from a well-initialized voxel encoder.

\begin{figure*}[t]
\centering
\includegraphics[width=17.5cm]{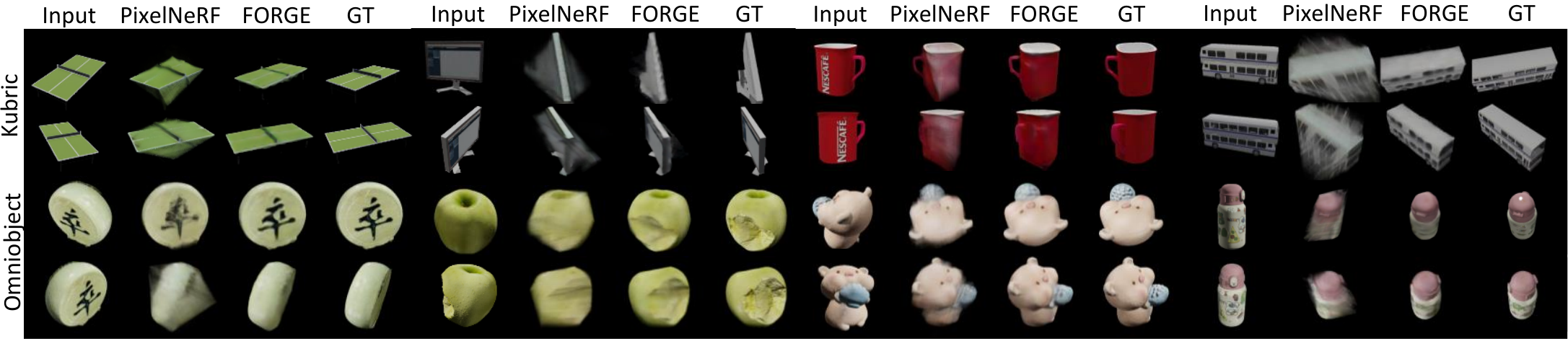}
\vspace{-0.26in}
\caption{\small{\textbf{Reconstruction results under ground-truth poses.} For each instance, we show two of the five inputs and two of the five novel views due to space limit. For the Kubric dataset, the left two and right two instances are from seen and novel categories. We observe that PixelNeRF usually collapses under large input image pose variation and is not robust to occlusion. FORGE handles these cases robustly.}
}
\vspace{-0.05in}
\label{fig: gt_compare}
\end{figure*}

\begin{table*}[t]
    \tiny
    \centering
    \tablestyle{5pt}{1.1}
    \vspace{-0.05in}
    \setlength\tabcolsep{3pt}
    \begin{tabular}{lcc|ccc|ccc|ccc|ccc}
    & & & \multicolumn{9}{c|}{GT Pose} &  \multicolumn{3}{c}{Pred. Pose} \\
    & & 
    & \multicolumn{3}{c}{PixelNeRF~\cite{Yu2021pixelNeRFNR}} 
    & \multicolumn{3}{c}{SRT~\cite{srt22}}
    & \multicolumn{3}{c|}{\modelname{}*}
    & \multicolumn{3}{c}{\modelname{}}\\
    
    \textit{Dataset} & \textit{Cat.} & \textit{Ins.} & 
    PSNR $\uparrow$ & SSIM $\uparrow$ & LPIPS $\downarrow$ &
    PSNR $\uparrow$ & SSIM $\uparrow$ & LPIPS $\downarrow$ &
    PSNR $\uparrow$ & SSIM $\uparrow$ & LPIPS $\downarrow$ &
    PSNR $\uparrow$ & SSIM $\uparrow$ & LPIPS $\downarrow$
     \\
    \shline
    \textit{Test1-seen} & \textit{13} & \textit{650} & 29.22 & 0.893 & 0.127 & 21.81 & 0.822  & 0.220 & \textbf{31.28} & \textbf{0.938} & \textbf{0.052} &  \textbf{26.61} & \textbf{0.896} & \textbf{0.106}\\ 
    \textit{Test1-unseen} & \textit{13} & \textit{650} & 29.27 & 0.894 & 0.125 & 21.86 & 0.820 & 0.213 & \textbf{31.36} & \textbf{0.938} & \textbf{0.053} & \textbf{26.62} & \textbf{0.895} & \textbf{0.105}\\\hline
    \textit{Test2-novel} & \textit{10} & \textit{1000} & 29.37 & 0.906 & 0.122 & 19.25 & 0.817& 0.236 & \textbf{31.17} & \textbf{0.946}& \textbf{0.058} & \textbf{25.57} & \textbf{0.898} & \textbf{0.107}
    \end{tabular}
    \vspace{-0.1in}
    \caption{\small{Evaluation of reconstruction quality on Kubric synthetic dataset. 
    We include FORGE$\dagger$ performance (w/o optimization) in ablation.}}
    \label{tabel: reconstruction}
\vspace{-0.15in}
\end{table*}

\begin{table}[t]
    \tiny
    \centering
    \tablestyle{4pt}{1.05}
    \vspace{-0.05in}
    \setlength\tabcolsep{5pt}
    \begin{tabular}{l|ccc|cc}
    & \multicolumn{3}{c|}{GT Pose} & \multicolumn{2}{c}{Pred. Pose} \\
    & P-NeRF~\cite{Yu2021pixelNeRFNR} & SRT~\cite{srt22} & FORGE* & FORGE$\dagger$ & FPRGE\\\shline
    Time & 3.1 min & 0.35 sec & \textbf{0.06 sec} & \textbf{0.09 sec} & 1.5 min
    \end{tabular}
    \vspace{-0.1in}
    \caption{\small{Inference time comparison.}}
    \vspace{-0.12in}
    \label{tabel: inference_time}
\vspace{-0.05in}
\end{table}

\section{Experiments}
\vspace{-0.05in}
As previous datasets~\cite{Sitzmann2019SceneRN, Xiang2014BeyondPA} contain a limited number of object categories or the images are captured under restricted camera settings, we design a new dataset to evaluate the performance of \modelname{}, especially for its generalization capability. We also test the performance of FORGE on public real object datasets, as well as apply it to in-the-wild images (in supplementary).

\vspace{-0.17in}
\paragraph{Kubric Synthetic Dataset.} We generate a new synthetic dataset using Kubric~\cite{Greff2022KubricAS}, which contains $23$ categories from ShapeNet~\cite{Chang2015ShapeNetAI}. For the training set, we randomly select $1000$ objects instances from each of the $13$ categories, having $13,000$ instances in total. We \textit{randomly} synthesize camera poses to render observations. 
The randomly sampled cameras make the dataset more realistic and challenging than the previous one that uses fixed and evenly distributed camera poses~\cite{Sitzmann2019SceneRN}.
We build \textbf{two test sets}. The first one (\textit{Test1}) contains $100$ instances for each of $13$ \textit{training categories} ($1300$ in total), where $50$ instances are \textit{seen} during training while with different rendering camera poses and $50$ are \textit{unseen}. 
The second one (\textit{Test2}) contains $100$ instances for each of $10$ \textit{novel categories} ($1000$ in total). We use $5$ views for input and another $5$ for evaluation.

\vspace{-0.17in}
\paragraph{OmniObject3D Dataset.} OmniObject3D~\cite{omniobject3d} is a large-scale real-object dataset. We randomly split its public content into $90\%$ (4901 instances) and $10\%$ (486 instances) for training and testing. The objects have rich realistic textures but simpler geometry compared with the Kubric dataset.

\vspace{-0.15in}
\paragraph{Google Scanned Object (GSO) Dataset.} We further test the generalization ability of \modelname{}, by applying the model \textit{trained on the Kubric synthetic dataset} to out-of-distribution real-objects from the GSO dataset~\cite{Downs2022GoogleSO}.
We include 300 samples from the GSO test set for evaluation, where all instances are from unseen categories. We show the results in supplementary due to the page limit.

\vspace{-0.17in}
\paragraph{Implementation Details.}
\label{sec: implimentation}
The resolution of inputs is $256\times 256$. We use $k=5$ images, separated into $2$ and $3$ views, to compute the cross-view consistency loss. We set $\lambda_{img}=5$, $\lambda_{p}=0.02$ and $\lambda_{pose}=1$.
The neural volume contains $64^3$ voxels. We sample $64$ points on each ray for rendering.
Please refer to the supplementary for more training details. We construct small subsets (about 50 instances) for Kurbic synthetic and OmniObject3D datasets for validation.

\vspace{-0.17in}
\paragraph{Metrics.} We adopt standard novel view synthesis metrics PSNR (in dB), SSIM~\cite{Wang2004ImageQA} and LPIPS to evaluate reconstruction results.
We also evaluate the relative camera pose error. We note that the pose translation error has different scales on the experimented datasets.

\begin{figure*}[t]
\centering
\includegraphics[width=17.5cm]{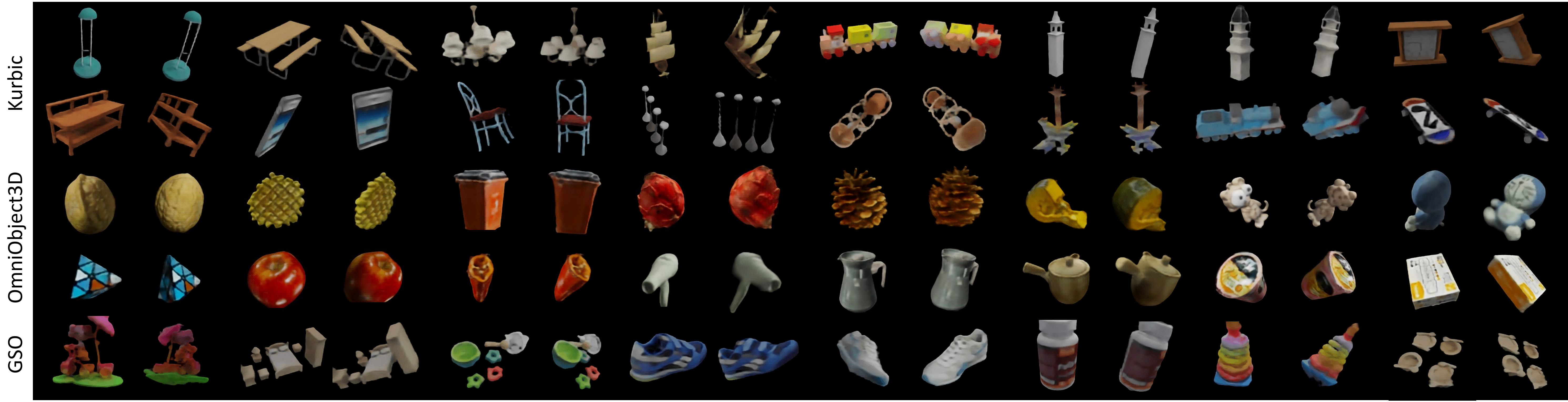}
\vspace{-0.25in}
\caption{\small{\textbf{Reconstruction results under predicted poses.} We visualize results on both the Kubric synthetic dataset, the OmniObject3D dataset, and the GSO dataset. We show two novel views of each instance.}
}
\vspace{-0.1in}
\label{fig: pred_compare}
\end{figure*}

\begin{figure*}[t]
\centering
\includegraphics[width=17.5cm]{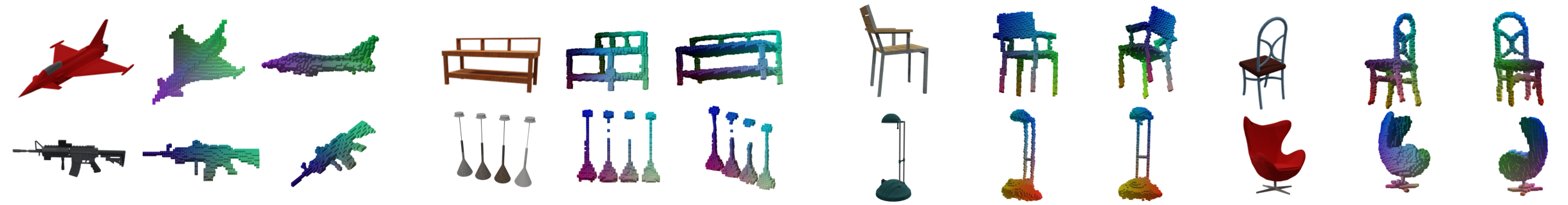}
\vspace{-0.25in}
\caption{\small{\textbf{Visualization of voxel reconstruction under predicted poses.} We obtain the voxel reconstruction by thresholding the predicted densities. Note that the thresholds are varied for different examples. We show the voxels in two views for each example.}
}
\label{fig: vis_voxel}
\end{figure*}

\begin{table*}[!t]
    \tiny
    \centering
    \tablestyle{6pt}{1.1}
    \vspace{-0.05in}
    \setlength\tabcolsep{5pt}
    \begin{tabular}{l|cc|cc|cc|cc|cc|cc}
    
    & \multicolumn{2}{c}{VideoAE~\cite{Lai2021VideoAS}} 
    & \multicolumn{2}{c}{8-Point TF\cite{Rockwell2022}}
    & \multicolumn{2}{c}{RelPose~\cite{zhang2022relpose}}
    & \multicolumn{2}{c}{Gen6D Refiner~\cite{Liu2022Gen6DGM}}
    & \multicolumn{2}{c|}{\modelname{}$\dagger$}
    & \multicolumn{2}{c}{\modelname{}}
    \\
    
    \textit{Dataset} & 
    \multicolumn{1}{c}{Rot. $\downarrow$}  & \multicolumn{1}{c}{Trans. $\downarrow$}
    & \multicolumn{1}{c}{Rot. $\downarrow$}  & \multicolumn{1}{c}{Trans. $\downarrow$}
    & \multicolumn{1}{c}{Rot. $\downarrow$}  & \multicolumn{1}{c}{Trans. $\downarrow$}
    & \multicolumn{1}{c}{Rot. $\downarrow$}  & \multicolumn{1}{c}{Trans. $\downarrow$}
    & \multicolumn{1}{c}{Rot. $\downarrow$}  & \multicolumn{1}{c|}{Trans. $\downarrow$}
    & \multicolumn{1}{c}{Rot. $\downarrow$}  & \multicolumn{1}{c}{Trans. $\downarrow$}
     \\
    \shline
    \textit{Test1-seen} & 36.7 & 0.56 & 15.0 & 0.32 & 24.7 & - & 17.2 & 0.36 & \textbf{7.3} & \textbf{0.17} & \textbf{4.0} & \textbf{0.11}\\ 
    \textit{Test1-unseen} & 33.8 & 0.57 & 15.4 & 0.32 & 24.0 & - & 17.3 & 0.35 & \textbf{7.9} & \textbf{0.18} & \textbf{4.3} & \textbf{0.11}\\\hline
    \textit{Test2-novel} & 42.7 & 0.69 & 23.9 & 0.49 & 32.3 & - & 22.8 & 0.52 & \textbf{13.7} & \textbf{0.31} & \textbf{8.9} & \textbf{0.23}\\
    \end{tabular}
    \vspace{-0.1in}
    \caption{\small{Evaluation of relative camera pose estimation on Kubric synthetic dataset. All models are trained with ground-truth poses.}}
    \label{tabel: pose}
\vspace{-0.13in}
\end{table*}

\vspace{-0.15in}
\paragraph{Our Tested Models}
\begin{itemize}
\item \modelname{}*. Our model uses ground-truth relative camera poses with the first view as the canonical view.
\item \modelname{}$\dagger$. Our model uses predicted relative poses with the first view as the canonical view.
\item \modelname{}. Our model with canonical view selection and quick test-time optimization (1K iterations) based on \modelname{}$\dagger$. We optimize the predicted poses using 2D photometric loss $L_{2D}$ (Sec.~\ref{sec: train}) on the input views. We perform $k$ times inference by setting every frame as canonical and use $L_{2D}$ as the selection criterion.
\vspace{-0.1in}
\end{itemize}

\begin{table}[t]
\scriptsize
\tablestyle{9pt}{1.05}
\setlength{\tabcolsep}{0.4em}
\begin{tabular}{l|c|cccc}
& Pose & PSNR $\uparrow$ & SSIM $\uparrow$  & Rot. error $\downarrow$ & Trans. error $\downarrow$ \\ \shline
PixelNeRF~\cite{Yu2021pixelNeRFNR} & GT & 26.97 & 0.888 & - & -\\
FORGE* & GT & \textbf{29.27} & \textbf{0.918} & - & - \\
8-Point TF~\cite{Rockwell2022} & Pred. & - & - & 13.69 & 0.72 \\
FORGE & Pred. & \textbf{26.56} & \textbf{0.889} & \textbf{5.35} & \textbf{0.30}\\
\end{tabular}
\vspace{-0.1in}
\caption{Results for both reconstruction and pose estimation on the Omniobject3D dataset.}
\vspace{-0.2in}
\label{table: omniobject}
\end{table}

\vspace{-0.1in}
\paragraph{Baselines.} For reconstruction, we compare with PixelNeRF~\cite{Yu2021pixelNeRFNR} and SRT~\cite{srt22} under ground-truth poses to validate the ability of \modelname{} on learning 3D geometry priors. For relative pose estimation, we compare with i) 2D-based VideoAE~\cite{Lai2021VideoAS}, 8-Point TF~\cite{Rockwell2022}, which separately predicts poses for each query view; ii) 2D-based RelPose~\cite{zhang2022relpose} which jointly estimate all relative poses, and iii) 3D-based Gen6D which separately predicts poses for each query view. All pose estimation baselines are trained with ground-truth poses. We use official or officially verified codes for baselines.

We alternatively notify that COLMAP~\cite{Schnberger2016StructurefromMotionR} can not be a baseline for comparison. The reason is that the Structure-from-Motion pipeline only works on videos or dense images (typically 50 images)~\cite{zhang2022relpose}. Its optimization process does not converge on 5 image inputs in most cases.


\begin{figure*}[t]
\centering
\includegraphics[width=\linewidth]{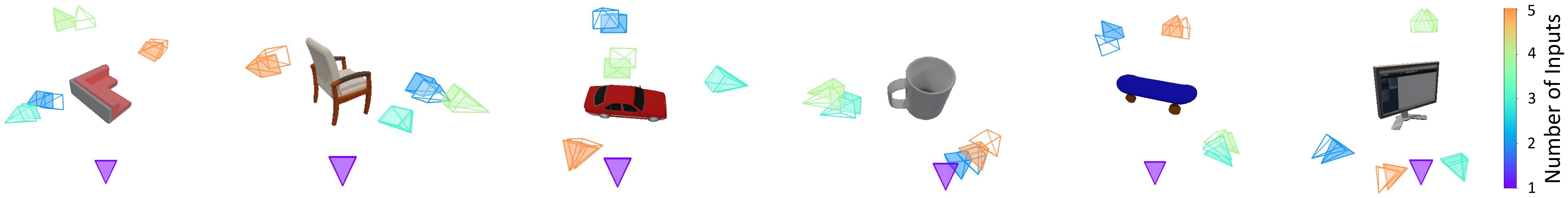}
\vspace{-0.25in}
\caption{\small{\textbf{Visualization of pose estimation.} Prediction and ground-truth are shown in the same colors with colored and white faces.}
}
\label{fig: vis_pose}
\end{figure*}

\begin{table*}[t]
    \tiny
    \centering
    \tablestyle{6pt}{1.1}
    \vspace{-0.05in}
    \begin{minipage}[t]{0.48\linewidth}
    \setlength\tabcolsep{4pt}
    \begin{tabular}{l|cc|cc|cc}
    
    & \multicolumn{2}{c}{w/ 3D U-Net~\cite{Zhou2018UNetAN}} 
    & \multicolumn{2}{c}{w/o CCL}
    & \multicolumn{2}{c}{\modelname{}*}\\
    
    \textit{Dataset} & \multicolumn{1}{c}{PSNR $\uparrow$}  & \multicolumn{1}{c}{SSIM $\uparrow$}
    & \multicolumn{1}{c}{PSNR $\uparrow$}  & \multicolumn{1}{c}{SSIM $\uparrow$}
    & \multicolumn{1}{c}{PSNR $\uparrow$}  & \multicolumn{1}{c}{SSIM $\uparrow$}
     \\
    \shline
    \textit{Test1} & \textbf{32.75} & \textbf{0.950} & 28.78 & 0.914 & 31.32 & 0.938\\ 
    \textit{Test2} & 22.33 & 0.826 & 28.05 & 0.918 & \textbf{31.17} & \textbf{0.946}\\
    \end{tabular}
    \vspace{-0.15in}
    \caption{\small{Ablation on voxel encoder design and the use of cross-view consistency loss (CCL). Experiments use G.T. poses.}}
    \label{tabel: ablation_encoder}
    \end{minipage}
    \hfill
    \begin{minipage}[t]{0.48\linewidth}
     \tablestyle{6pt}{1.1}
    \setlength\tabcolsep{4pt}
    \begin{tabular}{l|cc|cc|cc}
    
    & \multicolumn{2}{c}{Global} 
    & \multicolumn{2}{c}{Pairwise}
    & \multicolumn{2}{c}{Global-Pairwise}\\
    
    \textit{Dataset} & \multicolumn{1}{c}{Rot. $\downarrow$}  & \multicolumn{1}{c}{Trans. $\downarrow$}
    & \multicolumn{1}{c}{Rot. $\downarrow$}  & \multicolumn{1}{c}{Trans. $\downarrow$}
    & \multicolumn{1}{c}{Rot. $\downarrow$}  & \multicolumn{1}{c}{Trans. $\downarrow$}
    \\
    \shline
    \textit{Test1} & 12.9 & 0.24 & 11.3 & 0.21 & \textbf{8.0} & \textbf{0.17}\\ 
    \textit{Test2} & 26.2 & 0.44 & 22.1 & 0.42 & \textbf{13.8} & \textbf{0.31}\\
    \end{tabular}
    \vspace{-0.15in}
    \caption{\small{Ablation on the two feature extractors of pose estimation module. The models are not jointly trained with the voxel encoder.}}
    \label{tabel: ablation_pose}
    \end{minipage}
\end{table*}

\begin{table*}[!t]
    \tiny
    \centering
    \tablestyle{6pt}{1.1}
    \vspace{-0.05in}
    \begin{minipage}[t]{0.48\linewidth}
    \setlength\tabcolsep{4pt}
    \begin{tabular}{l|cc|cc|cc}
    & \multicolumn{2}{c}{Global w/o JR} 
    & \multicolumn{2}{c}{Pairwise w/o CCL}
    & \multicolumn{2}{c}{Pairwise-implicit}\\
    
    \textit{Dataset} & \multicolumn{1}{c}{Rot. $\downarrow$}  & \multicolumn{1}{c}{Trans. $\downarrow$}
    & \multicolumn{1}{c}{Rot. $\downarrow$}  & \multicolumn{1}{c}{Trans. $\downarrow$}
    & \multicolumn{1}{c}{Rot. $\downarrow$}  & \multicolumn{1}{c}{Trans. $\downarrow$}
     \\
    \shline
    \textit{Test1} & 15.1 & 0.28 & 20.1 & 0.33 & 23.4 & 0.37\\ 
    \textit{Test2} & 30.2 & 0.52 & 34.7 & 0.60 & 38.5 & 0.66\\
    \end{tabular}
    \vspace{-0.15in}
    \caption{\small{Ablation on the designs of two pose feature extractors.}}
    \label{tabel: ablation_pose2}
    \end{minipage}
    \hfill
    \begin{minipage}[t]{0.48\linewidth}
     \tablestyle{6pt}{1.1}
    \setlength\tabcolsep{4pt}
    \begin{tabular}{l|cc|cc|cc}
    
    & \multicolumn{2}{c}{only Avg. Fusion} 
    & \multicolumn{2}{c}{only Seq. Fusion}
    & \multicolumn{2}{c}{both (\modelname{}$\dagger$)}\\
    
    \textit{Dataset} & \multicolumn{1}{c}{PSNR $\uparrow$}  & \multicolumn{1}{c}{SSIM $\uparrow$}
    & \multicolumn{1}{c}{PSNR $\uparrow$}  & \multicolumn{1}{c}{SSIM $\uparrow$}
    & \multicolumn{1}{c}{PSNR $\uparrow$}  & \multicolumn{1}{c}{SSIM $\uparrow$}
     \\
    \shline
    \textit{Test1} & 24.23 & 0.852 & 25.36 & 0.864 & 26.08 & 0.879\\ 
    \textit{Test2} & 23.48 & 0.849 & 24.80 & 0.860 & 25.11 & 0.880 \\
    \end{tabular}
    \vspace{-0.15in}
    \caption{\small{Ablation on fusion methods and test-time optimization.}}
    \label{tabel: ablation_fusion}
    \end{minipage}
\vspace{-0.2in}
\end{table*}

\subsection{Reconstruction Evaluation}
\vspace{-0.05in}
\paragraph{Qualitative Results under Ground-Truth Pose.}

As shown in Fig.~\ref{fig: gt_compare}, we compare \modelname{} with PixelNeRF~\cite{Yu2021pixelNeRFNR} on both Kubric synthetic dataset and OmniObject3D dataset. We observe that results of PixelNeRF~\cite{Yu2021pixelNeRFNR} have strong artifacts when the input and target views have dramatic pose changes. PixelNeRF~\cite{Yu2021pixelNeRFNR} is also vulnerable under occlusion, while results of \modelname{} are more consistent. We conjecture that PixelNeRF~\cite{Yu2021pixelNeRFNR} is built on 2D representations, which cannot handle 2D ambiguity well. In comparison, \modelname{} benefits from our 3D-aware designs.

\vspace{-0.15in}
\paragraph{Qualitative Results under Predicted Pose.} 
We show the results in Fig.~\ref{fig: pred_compare}. \modelname{} can reconstruct the objects under noisy predicted poses. The geometry reconstruction performance on delicate objects and novel object categories match the results using ground truth poses. Besides, we visualize the voxel reconstruction based on the predicted density volume in Fig.~\ref{fig: vis_voxel}. It reveals that \modelname{} can capture 3D geometry accurately. Besides, we also include depth rendering error analysis in the supplementary.


\vspace{-0.17in}
\paragraph{Comparison with State-of-the-Art.}
As shown in Table~\ref{tabel: reconstruction}, when using ground truth poses, \modelname{} outperforms PixelNeRF~\cite{Yu2021pixelNeRFNR} and SRT~\cite{srt22} by a significant margin on both seen and novel object categories of Kubric synthetic dataset. Furthermore, the inference speed of \modelname{} is much faster than PixelNeRF, as shown in Table~\ref{tabel: inference_time}.
When using predicted poses, \modelname{} matches the performance using ground truth poses (\modelname{}*). The performance gap between seen and novel object categories is small, showing the strong generalization ability of \modelname{}. On the OmniObject3D dataset, we observe similar performance, as shown in Table~\ref{table: omniobject}. Moreover, the performance of \modelname{} using its predicted poses is close to the performance of PixelNeRF using ground-truth pose. This demonstrates the strong capability of \modelname{} on modeling real objects.

\subsection{Pose Estimation Evaluation}
\paragraph{Qualitative Results.} We visualize pose estimation results in Fig.~\ref{fig: vis_pose}. Our prediction matches ground truth accurately.

\vspace{-0.15in}
\paragraph{Comparison with State-of-the-Art.}
As Table~\ref{tabel: pose} shows, \modelname{}$\dagger$ achieves remarkable gain over previous methods. Both rotation and translation error is half of the best previous method 8-Point Transformer~\cite{Rockwell2022}. After test-time optimization, \modelname{} achieves $30\%$ performance gain. Moreover, our method shows a much smaller generalization gap, e.g., rotation error gap $5$ degree of \modelname{} vs. $9$ degree of \cite{Rockwell2022} on seen and novel object categories. We observe similar performance on the OmniObject3D dataset, as shown in Table~\ref{table: omniobject}, where \modelname{} decreases the pose error significantly.

\subsection{Ablation Study}
\vspace{-0.05in}
In this section, we verify the effectiveness of each proposed module on the Kubric synthetic dataset.

\vspace{-0.17in}
\paragraph{Voxel Encoder Design.}
We validate our design of the voxel encoder by comparing it with a counterpart using an additional 3D U-Net for feature refinement. As shown in Table~\ref{tabel: ablation_encoder}, using more 3D convolutions improves performance on seen object categories. However, the performance drops dramatically on novel object categories. This phenomenon implies that 3D U-Net overfits the training categories, which weakens its generalization ability. Instead, our encoder design shows great power in learning generalizable 3D priors.

\vspace{-0.15in}
\paragraph{Cross-view Consistency Loss.}
We study the impact of the cross-view consistency loss (CCL) to train the voxel encoder. As shown in Table~\ref{tabel: ablation_encoder}, the PSNR and SSIM drop $2.5$ dB and $0.25$, respectively, without using the CCL. 

\vspace{-0.15in}
\paragraph{Pose Estimator Design.} As Table~\ref{tabel: ablation_pose} shows, using either the global or pairwise pose feature extractor to predict poses achieves better performance than previous works in Table~\ref{tabel: pose}. Using both contribute to a stronger relative pose estimator.

\vspace{-0.15in}
\paragraph{Joint Pose Reasoning.}
We validate the effectiveness of performing joint pose reasoning (JR) using the self-attention blocks for the global pose feature extractor. For each global pose reasoning (GPR) module in the global pose feature extractor, we remove the self-attention block and double the cross-attention block, thus retaining the same model capacity. As shown in Table~\ref{tabel: ablation_pose2}, without joint pose reasoning, the performance of the global pose estimator drops $15\%$ compared to Global in Table~\ref{tabel: ablation_pose}. 

\vspace{-0.15in}
\paragraph{Cross-view Consistency Loss for Accurate Correlation.}
For the pairwise pose features extractor, we first study the importance of using CCL to build 3D correlation. As shown in Table~\ref{tabel: ablation_pose2}, when using the encoder trained without CCL, the performance drops $60\%$ compared to Pairwise in Table~\ref{tabel: ablation_pose}. It reveals the importance of enforcing cross-view consistency for finding reliable correlations.

\vspace{-0.15in}
\paragraph{Explicit 3D Correlation.}
We also verify that leveraging explicit 3D correlation is helpful for relative pose estimation. We compare the pairwise pose feature extractor with a counterpart that implicitly finds the correlation, denoted as Pairwise-implicit. For Pairwise-implicit, the value of attention is the canonical view features rather than the 3D positional encoding. As shown in Table~\ref{tabel: ablation_pose2}, the performance drops dramatically compared to the original model Pairwise in Table~\ref{tabel: ablation_pose}. We conjecture that finding explicit correlation helps 
mapping 3D voxel features to another feature space composed by the positional embedding. The feature space is consistent across different inputs, making the learning of pose regression easier.

\vspace{-0.15in}
\paragraph{Fusion Module Design.} We validate our designs on the feature fusion module by comparing the full fusion model with counterparts using (i) only average pooling fusion, and (ii) only sequential fusion. As shown in Table~\ref{tabel: ablation_fusion}, using average view-pooling fused features as initialization improves the sequential fusion.



\vspace{-0.05in}
\section{Conclusion}
We study the problem of reconstructing objects from a few views, where both the object categories information and camera poses are unknown.
Our key insight is leveraging the synergies between shape reconstruction and pose estimation to improve the performance of both tasks. We design a voxel encoder and a relative camera pose estimator, which are trained in a cascade manner. The pose estimator jointly reasons the poses of all views and establishes explicit 3D cross-view correlations. Then the 3D voxel features extracted from each view are transformed into a shared reconstruction space using the predicted poses. Then a neural volume is predicted, fusing per-view information.
Our model outperforms prior art on reconstruction and relative pose estimation by a significant margin. Ablation studies also show the effectiveness of each proposed module. We hope that our work inspires future efforts in making object reconstruction applicable and scalable in the real world.

\scriptsize{\textbf{Acknowledgement.} This work has been partially supported by
NSF CNS-1955523 and FRR-2145283, the MLL Research
Award from the Machine Learning Laboratory at UT-Austin,
and the Amazon Research Awards.}

{\small
\bibliographystyle{ieee_fullname}
\bibliography{egbib}
}

\appendix
\clearpage
\appendixpage
\section{Implementation Details}
In this section, we introduce the details of our model \modelname{}, including the voxel encoder, relative camera pose estimator, across-view feature fusion module, and volume rendering techniques. 

\subsection{Baselines}
We use official implementation for baselines and try our best to tune the parameters. For SRT~\cite{srt22}, we use an officially verified third-party implementation, as the official code is not released.

\subsection{Voxel Encoder}
The input of the 3D encoder is the RGB image observation and the output is per-view 3D features. The architecture is shown in Table~\ref{table: encoder3d arch}. For each 3D convolution layer, we use batch normalization with leaky ReLU activation.

\begin{table}[ht]
\scriptsize
\tablestyle{7pt}{1.2}
\begin{tabular}{ccc}\shline
 Stage & Configuration & Output\\\shline
 0 & Input image & $256\times 256 \times 3$\\\shline
  & {\textbf{2D Feature extraction}}  & \\ \hline
 \multirow{1}*{1} & \multirow{1}*{Res-50 (stride $8$)~\cite{He2016DeepRL}} & $32\times 32 \times 2048$\\ \shline
 & {\textbf{2D to 3D Convertion}}  & \\ \hline
 \multirow{1}*{2} & Reshape & $32\times 32 \times 32 \times 64$\\ \shline
 & {\textbf{3D Feature Processing}}  & \\ \hline
 \multirow{1}*{3} & Conv3D & $32\times 32 \times 32 \times 128$\\ \shline
\end{tabular}
\vspace{0.15in}
\vspace{-0.1in}
\caption{Network architecture of the voxel encoder.}
\label{table: encoder3d arch}
\end{table}

\subsection{Relative Camera Poses Estimator}
The relative camera poses estimator is composed by two feature extractor and a pose regression head. Their architecture are shown below.

\paragraph{Global Pose Feature Extractor.} The Global Pose Feature Extractor infers on 2D features of all inputs, which jointly reasons pose features of all views. The architecture is shown in Table~\ref{table: global pose}. In detail, the positional embedding added to each query view features are the same.

\begin{table}[t]
\scriptsize
\tablestyle{7pt}{1.2}
\begin{tabular}{ccc}\shline
 Stage & Configuration & Output\\\shline
 0 & Input image ($N$ views) & $k\times 256\times 256 \times 3$\\\shline
  & {\textbf{2D Feature extraction}}  & \\ \hline
 \multirow{1}*{1} & \multirow{1}*{FPN-P4~\cite{Lin2017FeaturePN}} & $k\times 16\times 16 \times 256$\\ \shline
 & {\textbf{Get Features}} & \\ \hline
 2 & Query view & $(k-1)\times 16\times 16 \times 256$\\ \hline
 2 & Canonical features & $ 1\times 16\times 16 \times 256$\\ \shline
 & {\textbf{Reshape}}  & \\ \hline
 3 & Query views & $[(k-1)\cdot 16\cdot 16] \times 256$\\ \hline
 3 & Canonical view & $[16\cdot 16] \times 256$\\ \shline
 & {\textbf{Add Pos. Embedding}}  & \\ \hline
 4 & Query views & $[(k-1)\cdot 16\cdot 16] \times 256$\\ \hline
 4 & Canonical view & $[16\cdot 16] \times 256$\\ \shline
 & \textbf{Global Pose Reasoning ($\times 3$)} & \\ \hline
 5 & Cross-attention & $[(k-1)\cdot 16\cdot 16] \times 256$\\ \hline
 6 & Self-attention &  $[(k-1)\cdot 16\cdot 16] \times 256$\\ \shline
 & \textbf{Reshape} & \\ \hline
 7 & Query views & $(k-1)\times 16\times 16 \times 256$\\ \shline
 & \textbf{Get Pose Features} & \\ \hline
 8 & 2D Convs & $(k-1)\times 1\times 1 \times 1024$  \\ \hline
 9 & Reshape & $(k-1) \times 1024$ \\ \shline
\end{tabular}
\vspace{0.15in}
\vspace{-0.1in}
\caption{Network architecture of the Global Pose Features Extractor.}
\label{table: global pose}
\end{table}

\paragraph{Pairwise Pose Features Extractor.} 
The Pairwise Pose Features Extractor infers on 3D features of inputs in a pairwise manner. The architecture is shown in Table~\ref{table: pairwise pose}.

\begin{table}[t]
\scriptsize
\tablestyle{7pt}{1.2}
\begin{tabular}{ccc}\shline
 Stage & Configuration & Output\\\shline
 0 & Query view features & $32\times 32 \times 32 \times 128$\\\hline
 0 & Canonical view features & $32\times 32 \times 32 \times 128$\\\hline
 0 & 3D Positional Encoding & $16\times 16 \times 16 \times 64$\\\shline
 & {\textbf{3D Feature Volume Processing}}  & \\ \hline
 1 & 2 Conv3D Layers & $16\times 16 \times 16 \times 64$\\\shline
 & {\textbf{Finding 3D correlation}}  & \\ \hline
 2 & Cross-view Transformer with P.E. & $16\times 16 \times 16 \times 64$\\\shline
 & {\textbf{correlation Refinement}}  & \\ \hline
 3 & Self-Attention & $16\times 16 \times 16 \times 64$\\\shline
 & {\textbf{Get pose features}}  & \\ \hline
 4 & Conv3D Layers & $1\times 1 \times 1 \times 1024$\\ \hline
 5 & Reshape & $1\times 1024$ \\
 \shline
\end{tabular}
\vspace{0.15in}
\vspace{-0.1in}
\caption{Network architecture and of the Pairwise Pose Features Extractor.}
\label{table: pairwise pose}
\end{table}

\paragraph{Pose Regression.} The architecture of the pose regression head is shown in Table~\ref{table: pose regression}.

\begin{table}[t]
\scriptsize
\tablestyle{7pt}{1.2}
\begin{tabular}{ccc}\shline
 Stage & Configuration & Output\\\shline
 0 & Features (Global) & $(k-1)\times 1024$\\\hline
 0 & Features (Pairwise) & $(k-1)\times 1024$\\\shline
 & \textbf{Merge Features} & \\ \hline
 1 & Concatenation & $(k-1)\times 2048$\\\shline
 & \textbf{Pose Regression} & \\ \hline
 2 & Dropout + MLP &  $(k-1)\times 7$\\\shline
\end{tabular}
\vspace{0.15in}
\vspace{-0.1in}
\caption{Network architecture and of the pose regression head.}
\label{table: pose regression}
\end{table}

\subsection{Feature Fusion Module}
The feature fusion module is composed by an average fusion block and a sequential fusion block. The inputs of the feature fusion module are the 3D features that have been transformed into the same reconstruction space, denoted as $\mathbb{\dot{Z}}_{3D} = \{\dot{z}_{3D}^i \vert i=1,...,k\}$.

First, we perform average pooling over the features as $Z_{3D} = Avg(\mathbb{\dot{Z}}_{3D})$, where $Avg(\cdot)$ is the average pooling operation for all views.

Then, we fuse the per-view 3D features to $Z_{3D}$ sequentially using ConvGRU. After we permute the sequence of 3D features by the distance between canonical view and the correspdoning view, the sequential fusion can be denoted as $Z_{3D} = GRU(Z_{3D}, \mathbb{\dot{Z}}_{3D})$, where $GRU$ is the ConvGRU, and $Z_{3D}$ serves as a feature initialization. Please refer to ~\cite{Ballas2016DelvingDI} for more details of sequential fusion using ConvGRU.


\begin{figure*}[t]
\centering
\includegraphics[width=\linewidth]{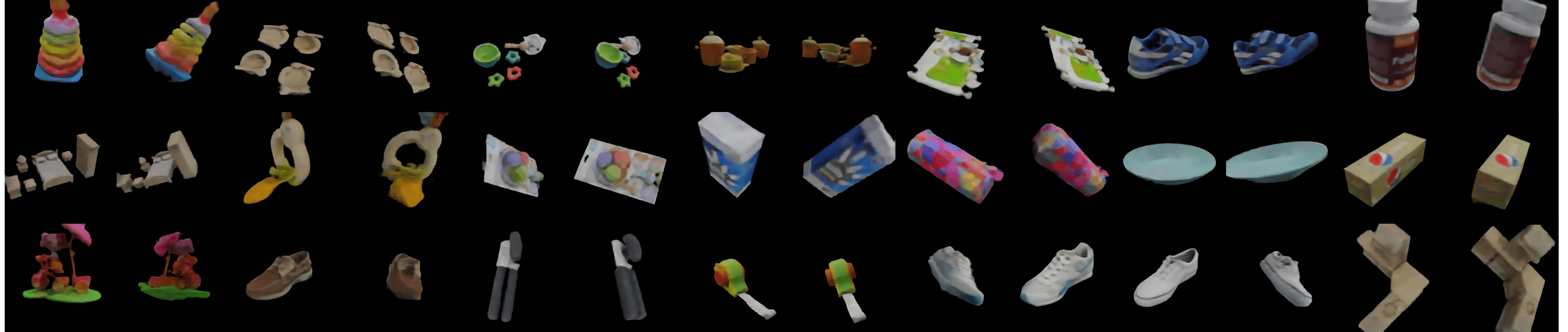}
\vspace{-0.25in}
\caption{Additional visualization on the Google Scanned Object (GSO) dataset~\cite{Downs2022GoogleSO}. The model is trained on the synthetic Kubric dataset and applied to GSO real objects directly. We show two novel views of the objects.
}
\label{fig: vis_gso}
\end{figure*}

\begin{figure*}[t]
\centering
\includegraphics[width=\linewidth]{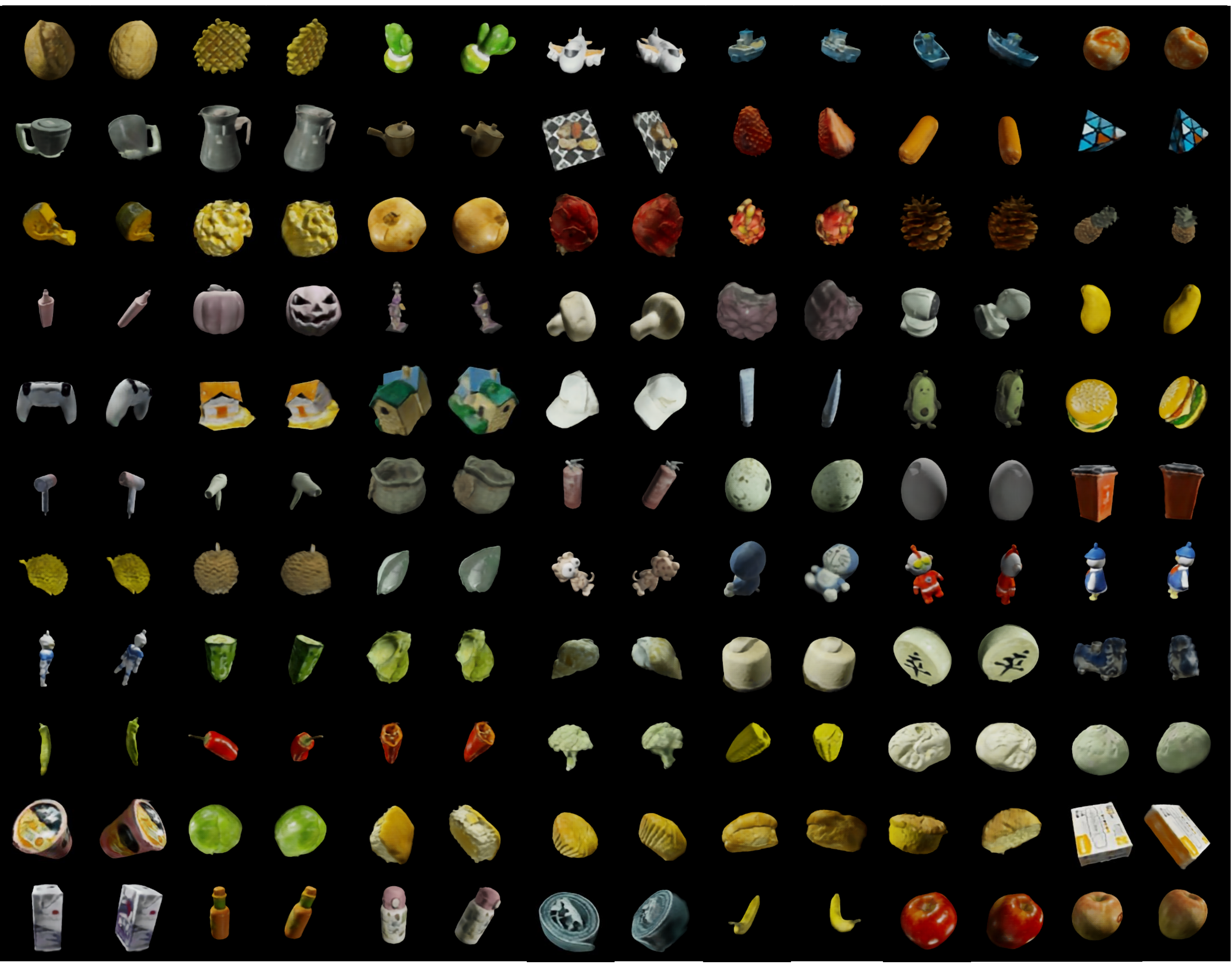}
\vspace{-0.25in}
\caption{Additional visualization on the OmniObject3D dataset~\cite{Downs2022GoogleSO}. The model is trained and tested on this dataset. We show two novel views of the objects.
}
\vspace{-0.1in}
\label{fig: vis_omniobject3d}
\end{figure*}

\subsection{Volume Rendering}
We perform parameterized volume rendering as render 3D features on 2D, and then perform 2D convolutions to get the final 2D reconstruction results. The input of the volume rendering is the 3D feature after cross-view fusion. We predict the neural volume $V \coloneqq (V_\sigma, V_f)$, where $V_\sigma, V_f$ are density volume and feature volume, from the 3D feature and use it for rendering. The architecture is shown in Table~\ref{table: volume rendering}.

\begin{table}[ht]
\scriptsize
\tablestyle{7pt}{1.2}
\begin{tabular}{ccc}\shline
 Stage & Configuration & Output\\\shline
 0 & 3D Feature & $32\times 32 \times 32 \times 128$\\\shline
 & \textbf{Predict Neural Volume} & \\ \hline
 1 & Density Volume & $64\times 64 \times 64 \times 1$\\\hline
 1 & Feature Volume & $64\times 64 \times 64 \times 16$\\\shline
  & {\textbf{Feature Volume Rendering}}  & \\ \hline
 \multirow{1}*{1} & \multirow{1}*{Volume Renderer} & $128\times 128 \times 16$\\ \shline
 & {\textbf{Feature Map to 2D Results}}  & \\ \hline
 \multirow{1}*{2} & Conv2D Layers & $256\times 256 \times 4$\\ \shline 
 & \textbf{Get RGB and mask} & \\ \hline
 3 & RGB & $256\times 256 \times 3$\\ \hline 3 & Mask & $256\times 256 \times 1$\\ \shline
\end{tabular}
\vspace{0.15in}
\vspace{-0.1in}
\caption{Network architecture and of the volume renderer.}
\label{table: volume rendering}
\end{table}

\section{Dataset}
We introduce more details of the Kubric Synthetic Dataset. To generate 2D observations, we randomly sample the cameras rotations. The camera are assigned with a distance to the object within a range of $[1.4, 1.6]$. we change the size of the objects by normalizing their axis-aligned bounding box to $1^3$ meter.

\paragraph{Training Categories.} We include 13 training categories, they are: \textit{airplane, bench, cabinet, car, chair, display, lamp, loudspeaker, rifle, sofa, table, telephone and vessel}. All of these 13 categories have more than $1000$ instances, ensuring the diversity of the datasets. We randomly sample $1000$ instances of each category for training.

\paragraph{Unseen Testing Categories.} We include 10 novel testing categories not seen during training, they are: \textit{bus, guitar, clock, bottle, train, mug, washer, skateboard, dishwasher, and pistol}. We randomly sample $100$ instances of each category for testing the generalization ability of \modelname{}.

\section{Training Details}
In this section, we introduce more details on training \modelname{}. We first train the model except the relative camera pose estimator. We use batch size $32$ with learning rate $0.0008$ for $40,000$ iterations. We half the learning rate at $15,000$, $30,000$ iterations. Then we train the relative camera pose estimator with fixing other parts of the model. We use batch size $40$ with learning rate $0.0002$ for $130,000$ iterations, where the learning rate is half at $50,000$, $70,000$, $90,000$ and $110,000$ iterations. Specifically, we train the Global Pose Features Extractor and Pairwise Pose Features Extractor separately and then finetune them together. In the end, we train the model with batch size $32$ and learning rate $0.0002$ for $30,000$ iterations, where the learning rate is half at $10,000$ and $20,000$ iterations. For validation, we use a small dataset with $50$ instances of training categories. We train our models with 8 A40 GPUs.

For volume rendering, The size of the predicted 3D volume is set to $1^3$ meter. For each camera frame, the volume is located at the center of the camera's frame, and the camera itself has a pose [$\mathbf{R}\vert \mathbf{t}$]. $\mathbf{R}$ is an identity matrix and $\mathbf{t}$ is set to $[0,0,1.5]$.


\section{More Results}
In this section, we provide more results on the real objects, including the Google Scanned Object (GSO) dataset~\cite{Downs2022GoogleSO} and the OmniObject3D dataset~\cite{omniobject3d}.

\subsection{Google Scanned Object Results}
We apply \modelname{} trained on Kubric synthetic dataset (which contains 13000 synthetic ShapeNet objects~\cite{Chang2015ShapeNetAI}), and test the generalization ability on novel real objects. 
We provide quantitative results in Table~\ref{table: gso}, where we compare with the best previous method PixelNeRF~\cite{Yu2021pixelNeRFNR} for novel view synthesis (verified by ~\cite{omniobject3d}), and compare with the best pose estimator 8-Point TF~\cite{Rockwell2022} to evaluate pose estimation performance. Our method \modelname{} demonstrates much better generalization ability on the out-of-domain real objects when using ground-truth poses. \modelname{} also achieves reasonable performance when using predicted poses, achieving comparable results with PixelNeRF with ground-truth poses. We visualize \modelname{} reconstructed objects in Fig.~\ref{fig: vis_gso}. The visualization demonstrates the strong generalization ability of \modelname{}.

\begin{figure}[t]
\centering
\includegraphics[width=1.0\linewidth]{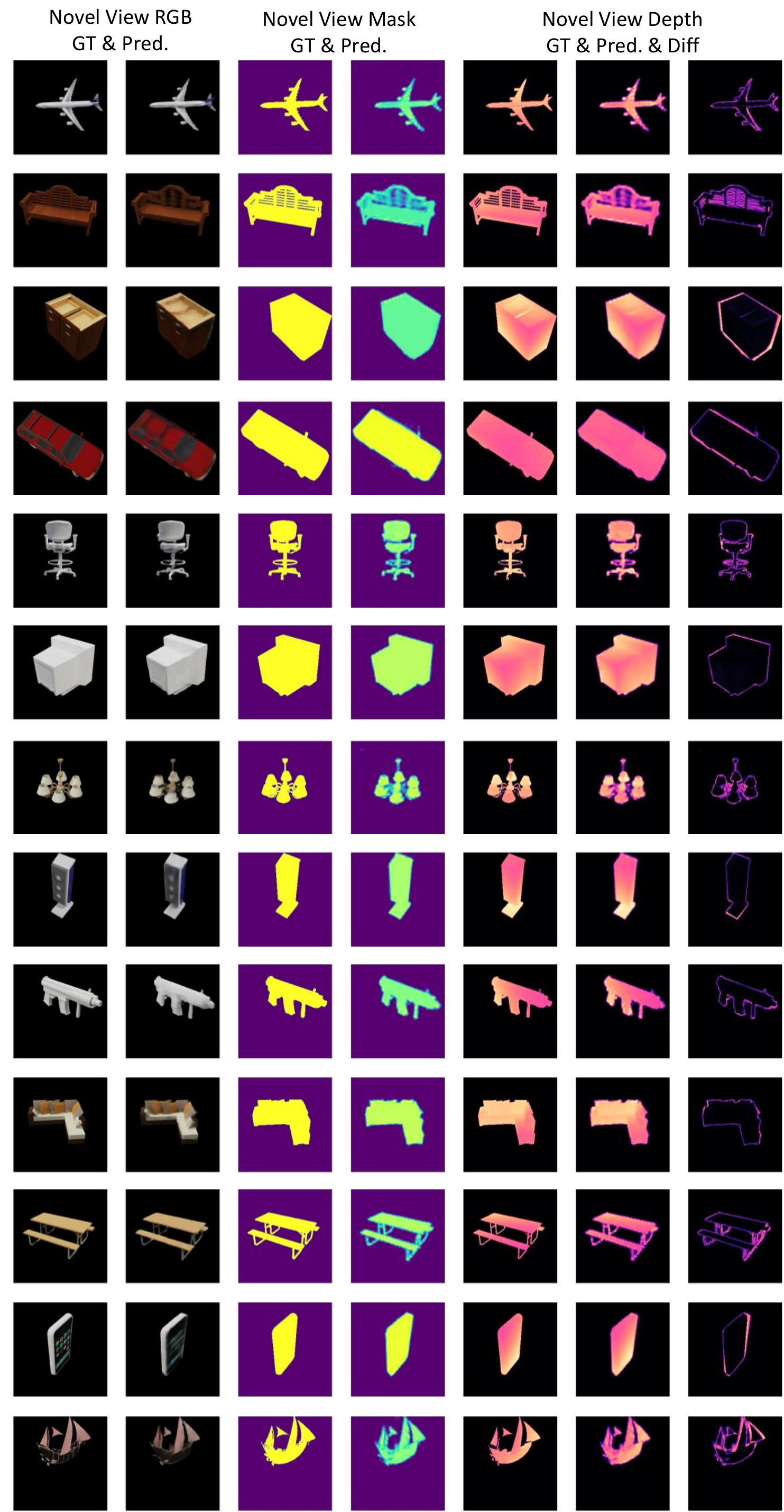}
\vspace{-0.25in}
\caption{\small{\textbf{Depth rednering visualization.} We show the GT depth, rendered depth, and their difference in the last three columns.}
}
\label{fig: depth}
\end{figure}

\begin{table}[t]
\vspace{-0.1in}
\scriptsize
\tablestyle{9pt}{1.05}
\setlength{\tabcolsep}{0.4em}
\begin{tabular}{l|c|cccc}
& Pose & PSNR $\uparrow$ & SSIM $\uparrow$  & Rot. error $\downarrow$ & Trans. error $\downarrow$ \\ \shline
PixelNeRF~\cite{Yu2021pixelNeRFNR} & G.T. & 23.27 & 0.789 & - & -\\
FORGE* & G.T. & \textbf{26.74} & \textbf{0.840} & - & - \\
8-Point TF~\cite{Rockwell2022} & Pred. & - & - & 22.63 & 0.60 \\
FORGE & Pred. & \textbf{22.78} & \textbf{0.772} & \textbf{14.90} & \textbf{0.37}\\
\end{tabular}
\vspace{-0.1in}
\caption{Results on GSO dataset. All methods are trained on the Kubric synthetic dataset.}
\vspace{-0.2in}
\label{table: gso}
\end{table}

\begin{table}[t]
\scriptsize
\tablestyle{9pt}{1.05}
\setlength{\tabcolsep}{0.4em}
\begin{tabular}{l|cc|c}
& \multicolumn{2}{c|}{GT Pose} & \multicolumn{1}{c}{Pred. Pose} \\
& PixelNeRF~\cite{Yu2021pixelNeRFNR} & FORGE* & FORGE\\ \shline
Seen & 0.130 & \textbf{0.085} & \textbf{0.160}\\
Novel & 0.123 & \textbf{0.082} & \textbf{0.158}\\
\end{tabular}
\vspace{-0.1in}
\caption{Depth rendering error on the Kubric synthetic dataset.}
\label{table: depth}
\end{table}


\begin{figure}[t]
\centering
\includegraphics[width=1.0\linewidth]{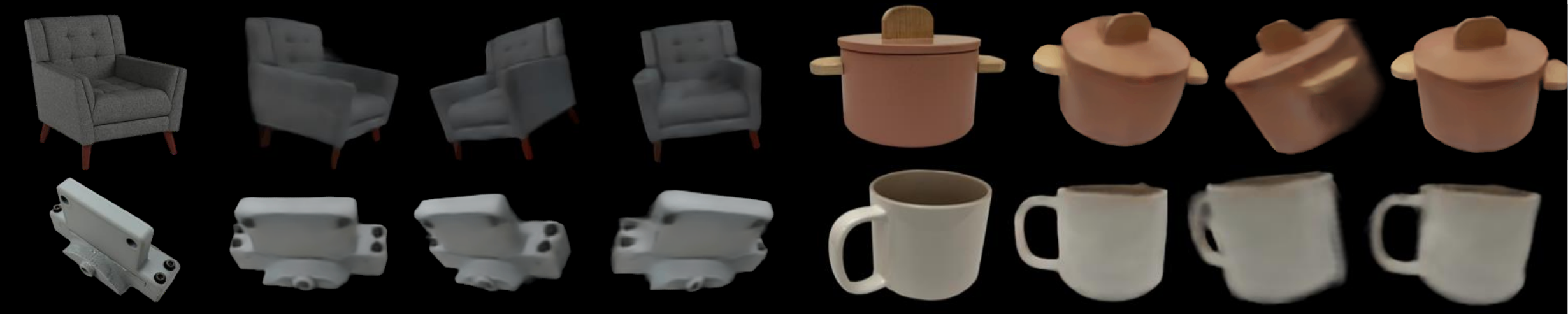}
\vspace{-0.25in}
\caption{\small{\textbf{In-the-wild reconstruction using images captured by iPhone or from Amazon products.} We show one input view and three novel views. All objects are from novel categories.}
}
\vspace{-0.05in}
\label{fig: real_demo}
\end{figure}

\begin{figure}[t]
\centering
\subfloat[]{
\label{fig: ablation_encoder}
\includegraphics[width=3.3cm]{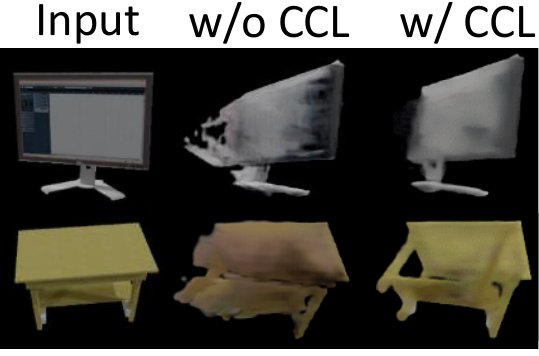}}
\hfill
\subfloat[]{
\label{fig: ablation_num_views}
\includegraphics[width=4.4cm]{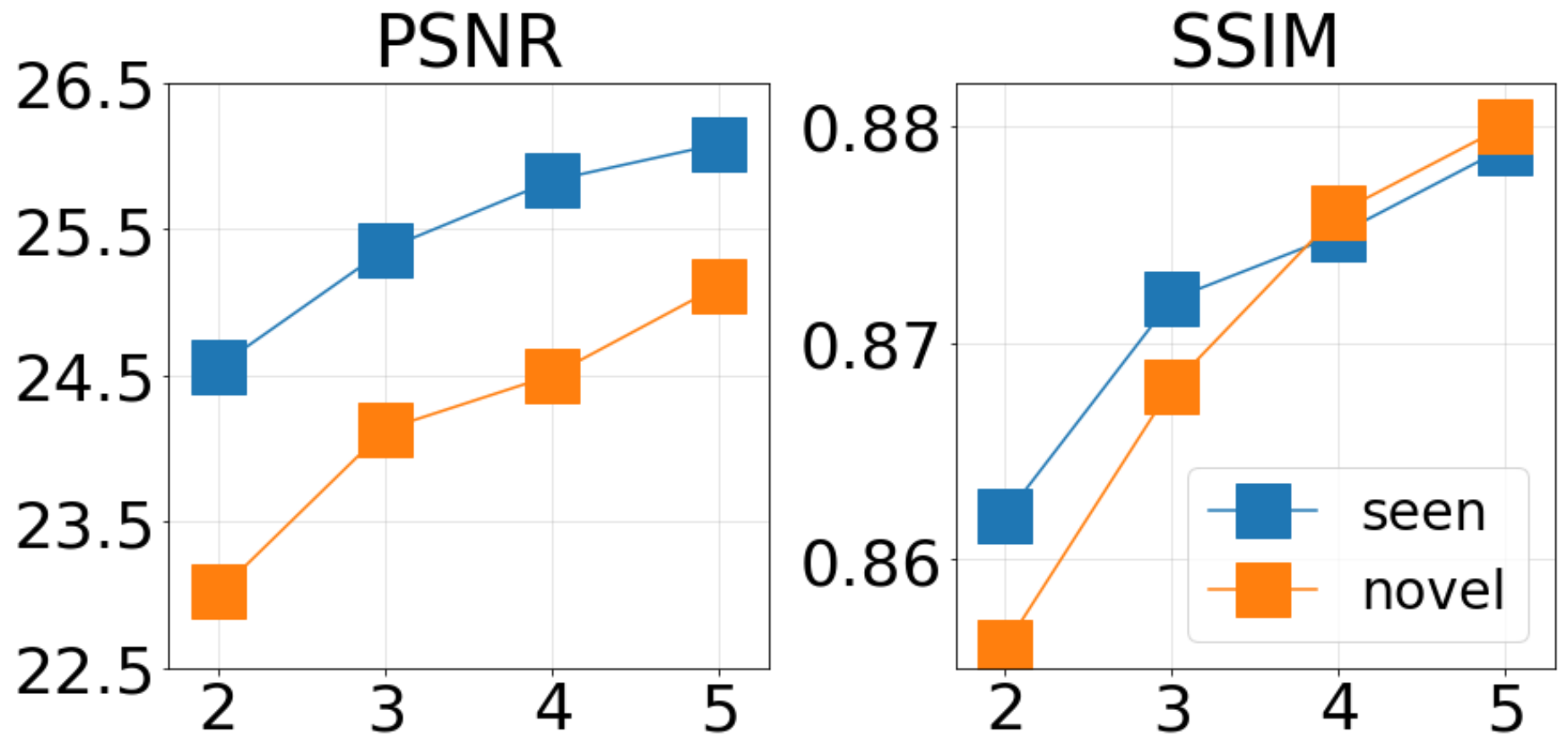}}
\vspace{-0.1in}
\caption{(a) \modelname{} reconstructs rarely visible object parts better with cross-view consistency loss (CCL); (b) Ablation on the number of input views on both seen and novel categories.}
\vspace{-0.15in}
\label{fig: ablation_merge_fig}
\end{figure}



\subsection{Real-life Object Reconstruction}.
We demonstrate that \modelname{} can reconstruct real-life objects from images in Fig.~\ref{fig: real_demo}. We use three views as input and an off-the-shelf method~\cite{Kirillov2020PointRendIS} to get object masks. Even though the domain gap exists, such as different camera intrinsics, \modelname{} can produce reasonable results.

\subsection{Object Geometry Accuracy}
In addition to the visualization of voxel reconstruction, to better evaluate the geometry of reconstructed objects, we also report the depth rendering error (L1 distance with depth ground-truth). We compare with PixelNeRF, as SRT is a 2D-based method and can not render depth.

\vspace{-0.15in}
\paragraph{Depth Reconstruction.}
As shown in Table.~\ref{table: depth}, when using ground-truth camera poses, FORGE demonstrates smaller depth rendering compared with PixelNeRF, showing its ability to model geometry accurately. When using predicted pose, the depth rendering error of \modelname{} is comparable to PixelNeRF which uses ground-truth pose. We visualize the rendered depth in Fig.~\ref{fig: depth}. We observe that the depth error of \modelname{} mainly originates from the boundary of the reconstructed object. This implies the accuracy of geometry modeling, while the pose error causes misalignment between ground-truth and rendered depth.

\subsection{In-the-wild Reconstruction} We include in-the-wild reconstruction examles in Fig.~\ref{fig: real_demo}.

\subsection{More Ablations}
In this section, we include more ablations to analyze the effectiveness of our proposed modules.
\vspace{-0.15in}
\paragraph{Cross-view Consistency Loss for Accurate Correlation.}
As visualization shown in Fig.~\ref{fig: ablation_encoder}, using CCL improves the reconstruction quality on 
object parts that are visible but heavily occluded.
It demonstrates that CCL regularizes the learning of 3D features by cross-view rendering. 

\vspace{-0.15in}
\paragraph{Number of Inputs.}
As shown in Fig.~\ref{fig: ablation_num_views}, we test the robustness of our model in handling varying numbers of input views from two to five. The performance is improved with more input views, and it tends to saturate with five.

\end{document}